\documentclass[runningheads]{llncs}

\usepackage{eccv}

\usepackage{eccvabbrv}

\usepackage{graphicx}
\usepackage{booktabs}
\usepackage{tabularx}
\usepackage{multirow}
\usepackage{authblk}

\usepackage[accsupp]{axessibility}  

\usepackage[pagebackref,breaklinks,colorlinks]{hyperref}

\usepackage{fancyhdr}
\begin{document}

\title{TFCounter: Polishing Gems for Training-Free Object Counting} 

\titlerunning{TFCounter}

\author{Pan Ting\inst{1} \textsuperscript{*}  \and
Jianfeng Lin\inst{1} \textsuperscript{*}\textsuperscript{†} \and
Wenhao Yu\inst{2} \and
Wenlong Zhang\inst{1} \and
Xiaoying Chen\inst{1} \and
Jinlu Zhang\inst{1} \and
Binqiang Huang\inst{1}
}

\renewcommand{\thefootnote}{\fnsymbol{footnote}} 
\footnotetext{\textsuperscript{*}Equal contributions}
\footnotetext{\textsuperscript{†}Corresponding Author: linjianfeng03@meituan.com}
\authorrunning{J.~Lin et al.}

\institute{Meituan \and
China University of Geosciences, Wuhan, China
}

\maketitle

\begin{abstract}
Object counting is a challenging task with broad application prospects in security surveillance, traffic management, and disease diagnosis. Existing object counting methods face a tri-fold challenge: achieving superior performance, maintaining high generalizability, and minimizing annotation costs. We develop a novel training-free class-agnostic object counter, TFCounter, which is prompt-context-aware via the cascade of the essential elements in large-scale foundation models. This approach employs an iterative counting framework with a dual prompt system to recognize a broader spectrum of objects varying in shape, appearance, and size. Besides, it introduces an innovative context-aware similarity module incorporating background context to enhance accuracy within messy scenes. To demonstrate cross-domain generalizability,  we collect a novel counting dataset named BIKE-1000, including exclusive 1000 images of shared bicycles from Meituan. Extensive experiments on FSC-147, CARPK, and BIKE-1000 datasets demonstrate that TFCounter outperforms existing leading training-free methods and exhibits competitive results compared to trained counterparts. Our code is available at \href{https://github.com/tfcounter/TFCounter}{https://github.com/tfcounter/TFCounter}
\end{abstract}

\section{Introduction}
Object counting, the task of estimating the number of specific objects within an image, plays a crucial role in various domains, including crowd counting\cite{liu2023point,liang2023crowdclip,abousamra2021localization,yang2022crowdformer,wang2020nwpu,zhang2016single,peng2018detecting,lian2019density,sindagi2019pushing,zhang2015cross} for urban planning and security, vehicle counting\cite{hsieh2017drone,mundhenk2016large} for traffic management, and cell counting\cite{tyagi2023degpr,wang2023cross,arteta2016detecting,xie2018microscopy} in medical applications. 

Traditional object-counting approaches are class-specific, counting objects belonging to predefined categories such as humans, cars, or cells. Typically grounded in CNN architectures, these methods require extensively annotated datasets. While exhibiting remarkable accuracy in dealing with trained categories, these methods fail to maintain their performance when counting novel classes during testing.
To address this limitation, recent researches\cite{ranjan2022exemplar,shi2022represent,yang2021class,ranjan2021learning,djukic2023low,lu2019class} have shifted towards class-agnostic object counting. They usually extract features from chosen exemplars and the query image to create a similarity map, which generates a density map to infer object count. This methodology
, exemplified in  \cite{djukic2023low}, allows for dynamic adaptation to arbitrary object classes, significantly broadening the scope and utility of object counting in computer vision.

\begin{figure}[t]
\centering
\includegraphics[height=7cm]{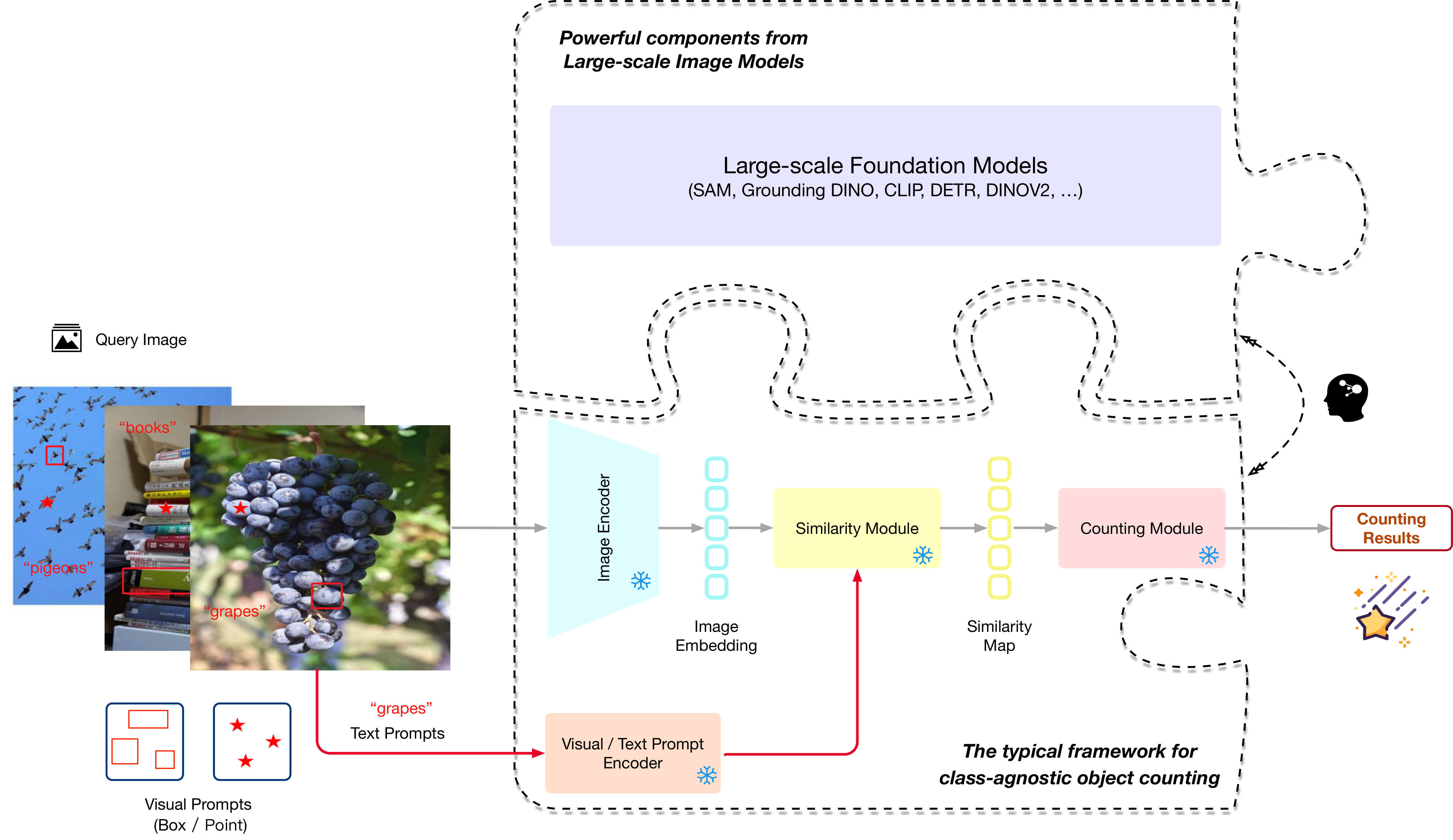}
\caption{Integrating task-specific frameworks with generalizable components from large-scale foundation models can achieve training-free class-agnostic object counting by detailed structural design.}
\label{fig:task-specific-frameworks}
\end{figure}

Recent progress in class-agnostic object counting has been directed towards two main objectives: improving the similarity map and minimizing labor annotations, including object points and bounding boxes, during testing and training.
(i) Numerous approaches\cite{ranjan2021learning,shi2022represent,yang2021class,you2023few,djukic2023low} have crafted specialized similarity module structures that primarily analyze foreground features. However, they tend to overlook background details, potentially compromising the precision of object counting.
(ii) Simultaneously, certain zero-shot models\cite{kang2023vlcounter,xu2023zero,amini2023open,jiang2023clip} utilize text prompts to identify object categories or count repeating classes in images, thus circumventing the requirement for box annotations in the testing phase. Yet, these models still require intensive dot annotations for each object during training, a challenging task in images with dense object clusters and frequent occlusions.
(iii) Additionally, the rapid advancement in large-scale foundation models\cite{kirillov2023segment,liu2023grounding,oquab2023dinov2,radford2021learning,carion2020end}, renowned for exceptional zero-shot generalization capabilities and flexibility in secondary development, has boosted interest in training-free approaches. A typical instance is the aligned text-image encoder in CLIP\cite{radford2021learning}, proving its adaptability across a wide range of downstream tasks\cite{jain2022zero,yan2023unloc,yang2023vid2seq,xu2023open}, achievable through parameter freezing or fine-tuning.
Leveraging these foundation models, some methods\cite{shi2023training,liu2023grounding} can perform training-free object counting by directly processing the output results or innovative structural designs, as shown in \ref{fig:task-specific-frameworks}. Nevertheless, these methods often trade-off between high performance and broad generalizability.

Driven by this analysis, we introduce TFCounter, as shown in Figure \ref{fig:TFCounter}, a novel training-free class-agnostic object counter, which is prompt-context-aware via the cascade of the essential elements in large-scale foundation models. This approach performs a multi-round counting strategy that utilizes posterior knowledge to broaden the recall scope. Subsequently, it introduces an innovative context-aware similarity module incorporating background context to enhance accuracy. Moreover, it uses two types of point prompts, matrix point prompt and residual point prompt, with the latter specifically designed to capture small objects that are often missed. This dual prompt system ensures comprehensive object detection across various sizes. Finally, to validate the effectiveness and generalizability of TFCounter, we introduce an exclusive dataset named BIKE-1000, comprising 1000 images of shared bicycles from Meituan. Experimental results show that TFCounter outperforms existing state-of-the-art training-free models on two standard counting benchmarks, and even displays competitive performance when compared with trained models. In short, our contributions can be summarized as follows:
\begin{itemize}
 \item We introduce TFCounter, a novel training-free class-agnostic object counter, prompt-context-aware via the cascade of the essential elements in large-scale foundation models. 
\item We propose a context-aware similarity module for improved precision and an iterative counting framework with a dual prompt system for broader recall.
\item We present a novel exclusive dataset named BIKE-1000 for object counting, which validates the superior performance of TFCounter.
\end{itemize}

\section{Related Works}
\subsubsection{Improved similarity map} approaches have attracted significant interest in class-agnostic object counting. Several models strived to generate high-quality similarity maps for refined count inference. FamNet+\cite{ranjan2021learning} introduced a novel adaptation strategy for few-shot regression counting, adapting the model to new visual categories at test time with a few exemplars. BMNet\cite{shi2022represent} and its extension, BMNet+\cite{shi2022represent}, focused on a similarity-aware framework with a learnable bilinear similarity metric. CFOCNet+\cite{yang2021class} used a two-stream Resnet for different scales similarity calculation and aggregation. SAFECount\cite{you2023few} proposed a learning block with a similarity comparison module and a feature enhancement module, while LOCA\cite{djukic2023low} developed an object prototype extraction module for low-shot counting problems. However, these methods often overlooked background considerations in favor of foreground focus.

\subsubsection{Minimizing labor annotations} is another focus in the class-agnostic object counting task. Existing methods frequently depended on annotations such as points and boxes during training and testing. To improve flexibility, several approaches aimed to eliminate human-annotated bounding boxes during testing, achieving zero-shot counting. Among these, EF-CAC\cite{ranjan2022exemplar} counted all repeating objects through the region proposal network, while ZSC\cite{xu2023zero}, CounTX\cite{amini2023open} and CLIP-Count\cite{jiang2023clip} accepted an arbitrary object class description to predict the object number. Concurrently, other methods were designed for training-free object counting, capitalizing on the robustness and generalizability inherent in large-scale foundational models. SAM\cite{kirillov2023segment} could perform zero-shot segmentation and subsequently estimated the number of objects by tallying all the generated masks. Based on it, SAM-Free\cite{shi2023training} combined three distinct types of class-specific priors to improve efficiency and accuracy. GroundingDINO\cite{liu2023grounding} exceled in open-set detection, capable of object counting via aggregation of detected bounding boxes. Nevertheless, zero-shot models often necessitated extensive point annotations during the training phase, which posed a challenge especially in complex scenes. Training-free methods typically struggled in complex scenes or exhibited constraints in their ability to generalize across multiple object categories.

\section{Methodology}
\subsection{Network Architecture}
\begin{figure}[t!]
\centering
\includegraphics[height=7.2cm]{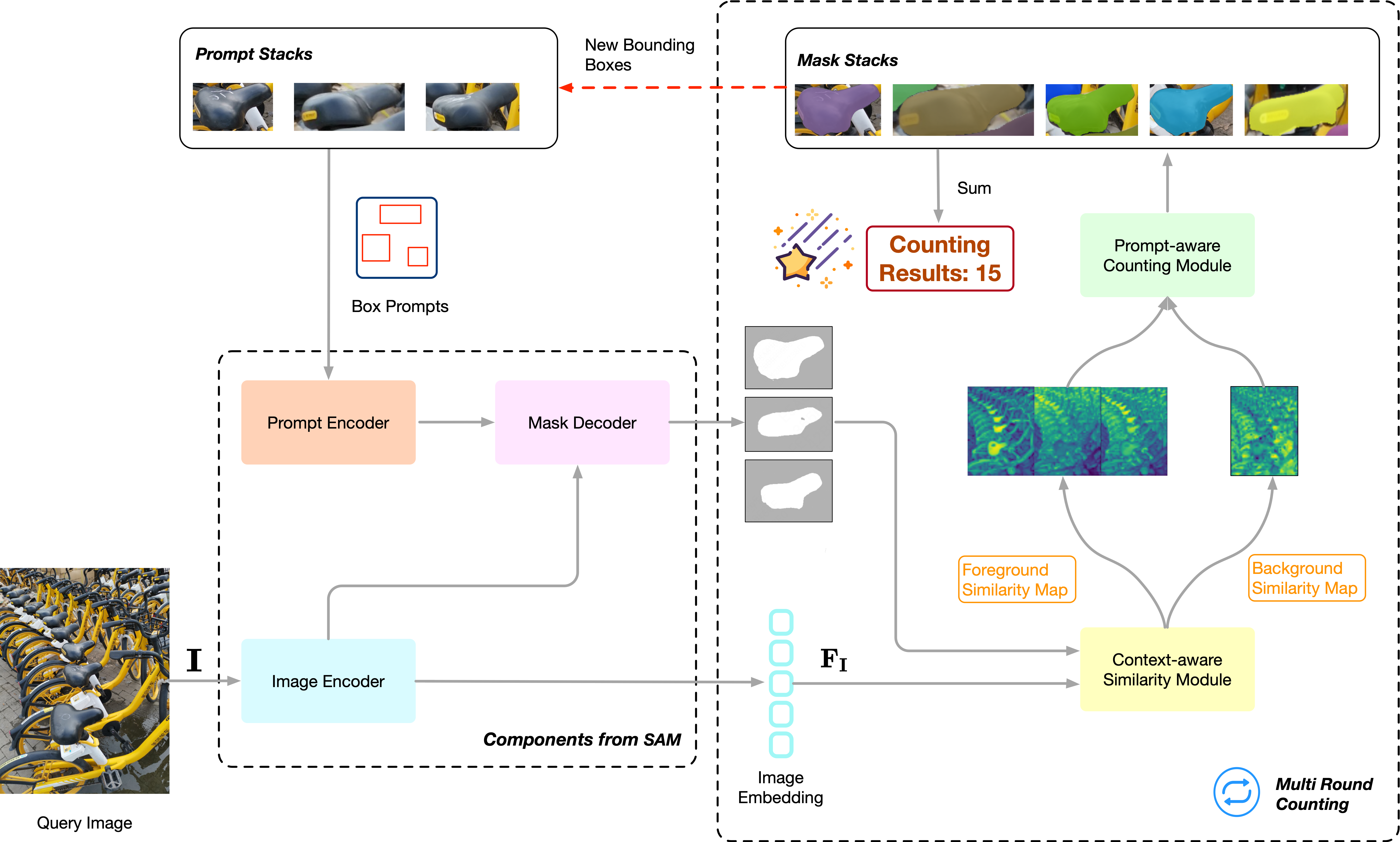}
\caption{Overview of our TFCounter. TFCounter is a segmentation-based model designed for training-free, class-agnostic object counting. It employs an iterative counting mechanism and links three key modules: feature encoding, context-aware similarity computation, and prompt-aware object counting.}
\label{fig:TFCounter}
\end{figure}

Class-agnostic object counting aims to enumerate exemplar objects in a query image with minimal support images. In this paper, we propose a novel training-free method, an overview of which is illustrated in Figure \ref{fig:TFCounter}. 

Our method starts with segmenting exemplars in the query image, utilizing SAM\cite{kirillov2023segment} as our backbone, to generate corresponding foreground masks. Following this, we introduce a novel context-aware similarity module considering both foreground and background context to compute similarity maps. Then, the prompt-aware counting module performs a dual prompt system on the weighted-fusion similarity map to generate masks for target objects. Central to our method is an iterative counting mechanism that compares the minimum bounding boxes generated from the mask stacks with those from the prompt stacks. This process is triggered upon detecting new bounding boxes until a set iteration cap is reached. This iterative strategy establishes an approximation chain, enhancing the model's generalization capabilities to recognize a broad spectrum of objects. Finally, we determine the number of objects by counting all masks in the mask stacks, thus concluding our object-counting process.

\subsection{Context-aware Similarity Module}
\begin{figure}[t!]
\centering
\includegraphics[height=9.2cm]{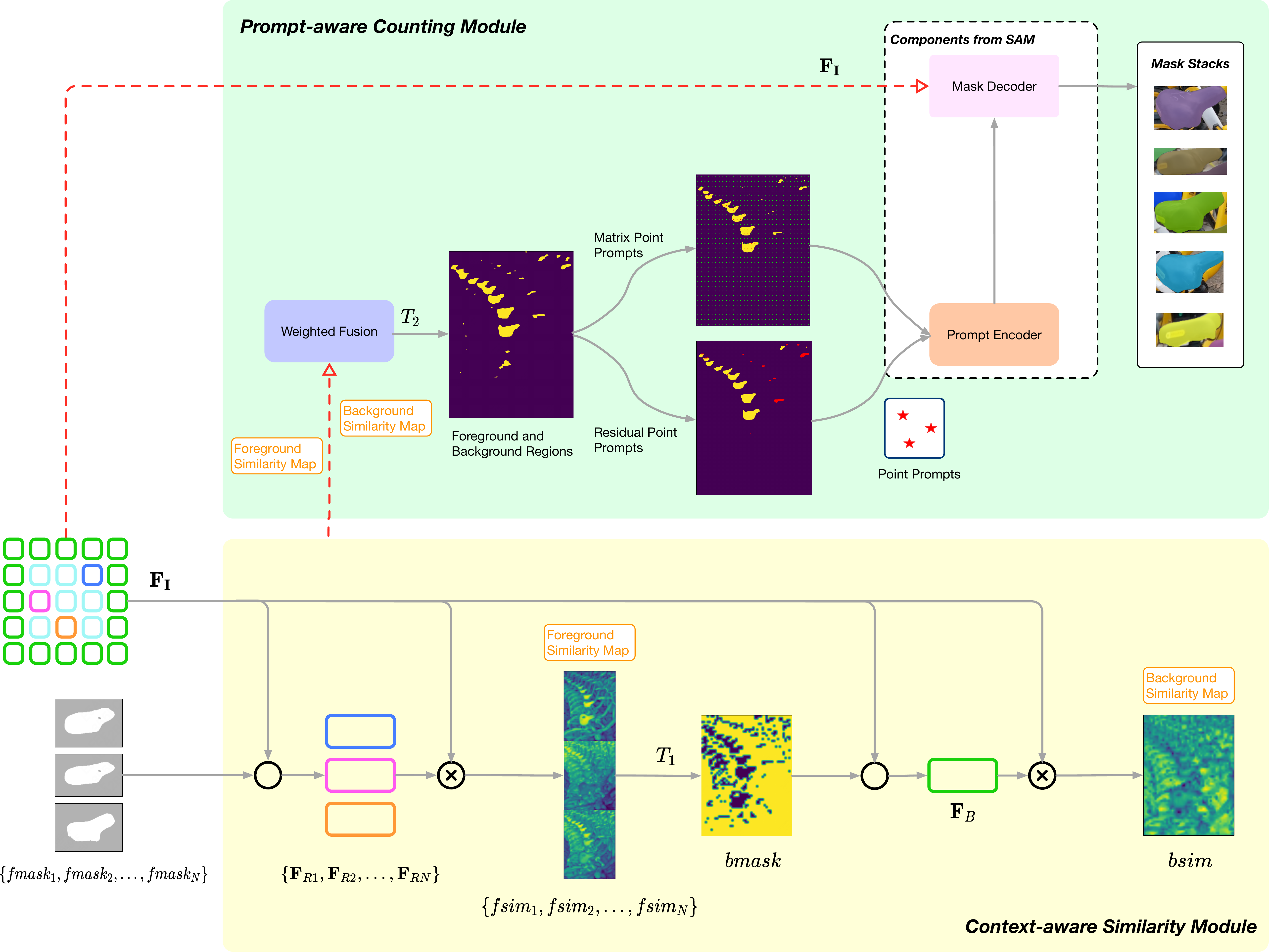}
\caption{The context-aware similarity module utilizes the image embedding $\mathbf{F} _{\mathbf{I}}$ and all foreground masks $\{fmask_1, fmask_2, ... , fmask_N\}$ to generate both foreground and background similarity maps. The prompt-aware counting module performs weighted fusion on these similarity maps and operates a dual prompt system to produce target masks.}
\label{fig:Similarity Module}
\end{figure}

The context-aware similarity module,  as shown in Figure \ref{fig:Similarity Module}, generates a series of foreground similarity maps and one background similarity map by processing the image embedding and the foreground masks. We define the image embedding $\mathbf{F} _{\mathbf{I} }$ and $i$-th foreground mask $fmask_i$. 

First, the module extracts $i$-th foreground feature embedding $\mathbf{F} _{Ri}$ through the Hadamard product, as indicated in Formula \ref{formula:{F} _{Ri}}:
\begin{equation}
    \mathbf{F} _{Ri} = fmask_i \circ \mathbf{F} _{\mathbf{I}}
    \label{formula:{F} _{Ri}}
\end{equation}
where $\circ$ denotes the Hadamard product, $x$ and $y$ represent spatial coordinates in $\mathbf{F} _{\mathbf{I}}$, and 
\begin{equation}
    \mathbf{F} _{Ri}[x,y] = fmask_i[x,y] * \mathbf{F} _{\mathbf{I}}[x,y]
\end{equation}
Then it computes the corresponding foreground similarity map $fsmi_i$ by performing a Euclidean dot product between $\mathbf{F} _{\mathbf{I} }$ and $\mathbf{F} _{Ri}$, as illustrated in Formula \ref{formula:fsim_i}: 
\begin{equation}
    fsim_i = \mathbf{F} _{Ri} \otimes \mathbf{F} _{\mathbf{I}}
    \label{formula:fsim_i}
\end{equation}
where $\otimes $ indicates the Euclidean dot product, $u$ and $v$ represent spatial coordinates in $\mathbf{F} _{Ri}$, $K$ denotes the number of non-zero pixels of $\mathbf{F} _{Ri}$, and 
\begin{equation}
fsim_i[x,y] = \frac{1}{K} \sum_{u,v}^{}\mathbf{F} _{\mathbf{I}}[x,y]*\mathbf{F} _{Ri}[u,v]
\end{equation}
According to this method, we can calculate all foreground similarity maps for $N$ prompt boxes, where pixel values represent the degree of correlation between each pixel in the query image and the corresponding exemplar. Subsequently, the module calculates the average of all foreground similarity maps for the background analysis. We use the binarization threshold $T_1$ to generate the background mask $bmask$. Finally, the background similarity map $bsim$ is produced in a manner analogous to the generation process of $fsmi_i$, thereby completing the similarity assessment process of our model.

\subsection{Prompt-aware Counting Module}
The prompt-aware counting module,  as indicated in Figure \ref{fig:Similarity Module}, combines all similarity maps with the image embedding to produce masks for target objects. 

First, the module creates a composite similarity map $csim$ by weighting and fusing whole foreground and background similarity maps, as shown in Formula \ref{formula:csim} 
\begin{equation}
    csim = \frac{1}{N} \sum_{i=1}^{N} fsim_{i} + \lambda \times bsim
    \label{formula:csim}
\end{equation}
where $\lambda$ is a key hyperparameter and $N$ denotes the number of prompt boxes. This fusion enhances the distinction between foreground and background regions for more accurate segmentation. Then, we segment these regions on the composite similarity map with the threshold $T_2$ and perform a dual prompt system to generate two types of point prompts: (1) matrix point prompt, which is set to 1 within the foreground and 0 otherwise, processed in batches for efficiency; (2) residual point prompt, which marks unmasked foreground areas as 1, targeting small objects that the matrix prompt may miss. Finally, the prompt encoder and the mask decoder transform these point prompts into target masks, which are then compiled in the mask stacks for subsequent analysis.

\section{BIKE-1000 Dataset}
\begin{figure}[t!]
\centering
\includegraphics[height=6.2cm]{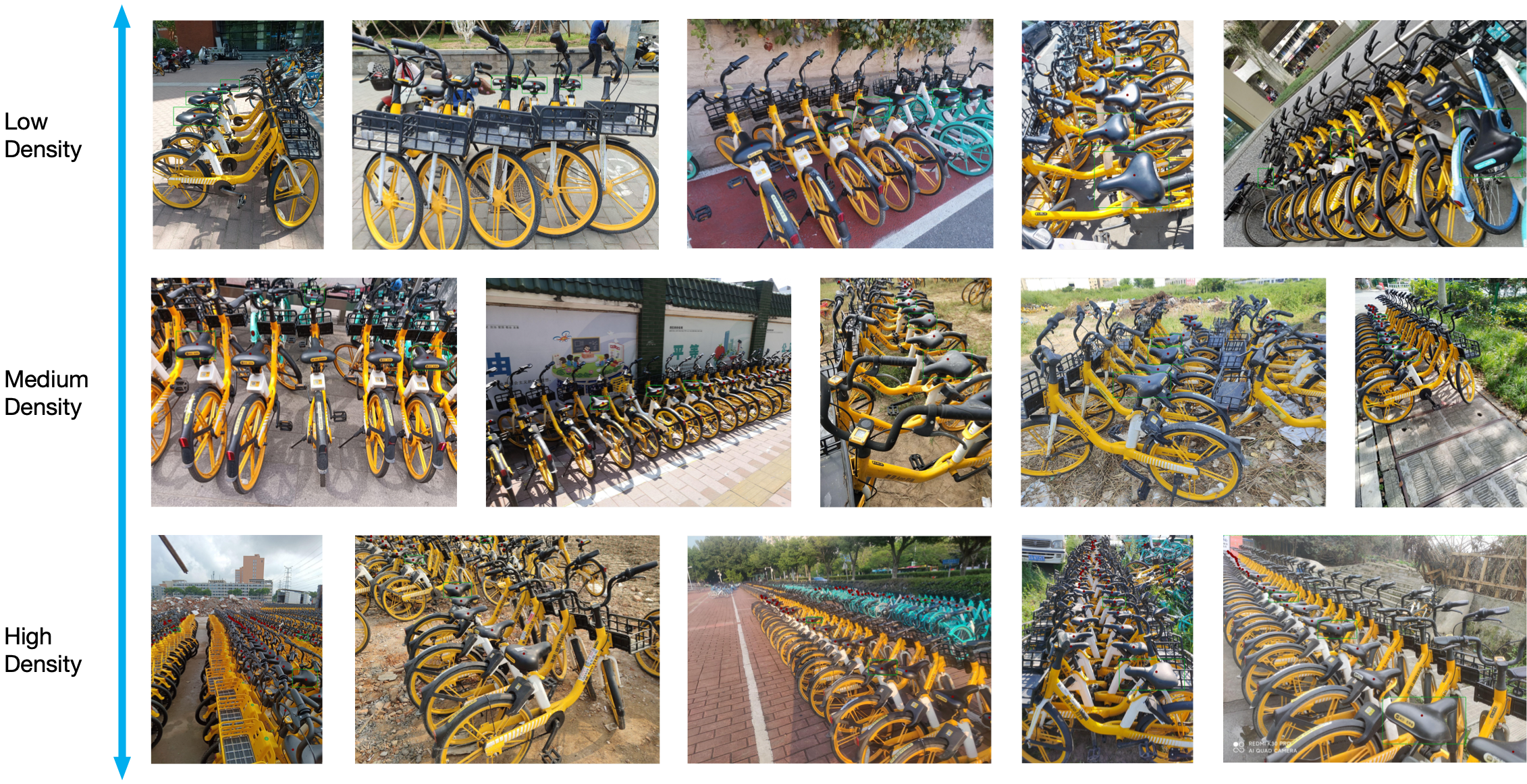}
\caption{Few annotated images from BIKE-1000. Dot and box annotations are indicated in red and green, respectively. Most images feature an oblique perspective, leading to bicycles with considerable variations in shapes, appearances, and sizes, even instances of occlusion.}
\label{fig:Few annotated images from the dataset}
\end{figure}

\begin{figure}[t!]
\begin{minipage}[t]{0.48\textwidth}
\centering
\vspace{0pt}
\includegraphics[width=\textwidth]{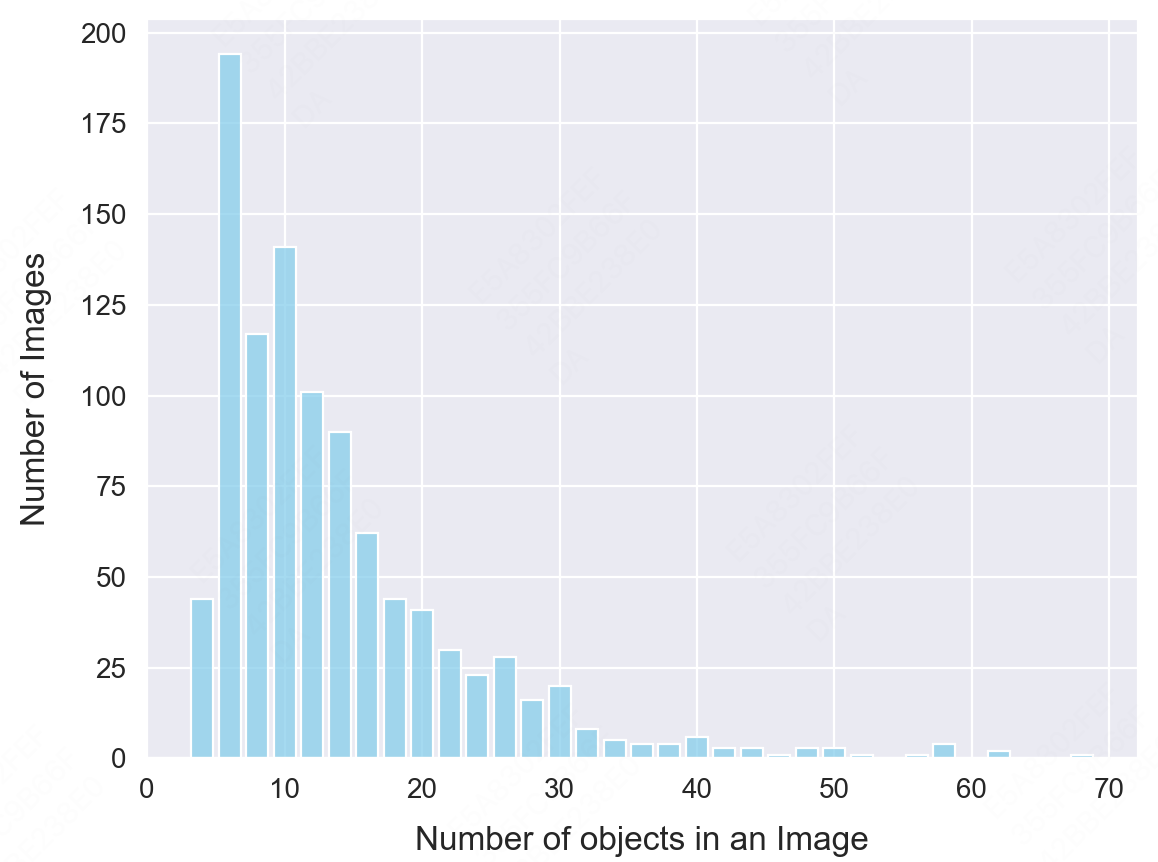}
\caption{Number of images in several ranges of object count.}
\label{fig:statistical data}
\end{minipage}
\hfill
\begin{minipage}[t]{0.48\textwidth}
\centering
\vspace{0pt}
\captionof{table}{Comparison with popular object counting datasets: "$v$" for vertical perspective, "$o$" for oblique; "$b$" for bounding box annotations, and "$p$" for point.}
\label{tab:Comparison with popular counting datasets}
\renewcommand{\arraystretch}{1.4} 
\scriptsize
\begin{tabularx}{\textwidth}{XXXX}
\hline
Dataset         & CARPK    & FSC147           & BIKE-1000 \\ \hline
Year            & 2017     & 2021             & 2023      \\
Images          & 1448     & 6135             & 1000      \\
Categories      & 1        & 147              & 1         \\
Instances       & 43       & 56               & 13        \\
Perspective     & $v$      & $v$.$o$          & $o$   \\
Annotation      & $b$.$p$  & $b$.$p$          & $b$.$p$   \\ \hline
\end{tabularx}
\end{minipage}
\end{figure}

This paper utilizes exclusive data from Meituan, one of China's leading shared bicycle enterprises. In the bike-sharing and ebike-sharing industry, accurate bicycle counting is a central requirement across multiple application scenarios, including orderly operations management, inventory audits, and street silt removal. To support these scenarios and advance the research and development of more precise and efficient counting technologies, we have established a novel object counting dataset named BIKE-1000. This dataset provides a large collection of bicycle images accompanied by their count annotations, which aids in improving bicycle management, enhancing operational efficiency, and ultimately optimizing the user experience.

The BIKE-1000 dataset encompasses a collection of 1000 images, each featuring distinctly visible shared bicycles situated within various scenes. These images were primarily captured by operators. A significant portion of the dataset is characterized by photographs taken from an oblique perspective, which presents the bicycles with considerable variations in shape, appearance, and size, as well as instances of partial occlusion. Such attributes pose typical challenges in the domain of object counting in computer vision. The annotation protocol for the BIKE-1000 dataset adheres to the methodology used in FSC147 \cite{ranjan2021learning}, comprising (1) point annotation, where each countable bicycle seat is marked, and (2) bounding box annotation, with three instances per image demarcated as examples. The dataset includes high-resolution imagery with bicycles ranging from 3 to 70 per image, averaging 13 objects. Note that shared bicycles consist of numerous components, such as frames, handlebars, wheels, seats, etc., whose appearance can vary significantly when viewed from different angles. Manually counting over 70 bicycle seats in a single image proved difficult, especially in images with oblique perspective. Therefore, we have limited our image selection to those with fewer than 70 bicycle seats for the BIKE-1000 dataset. The visualizations are displayed in Figure \ref{fig:Few annotated images from the dataset}, while the statistical data and comparisons with object count benchmarks are shown in Figure \ref{fig:statistical data} and Table \ref{tab:Comparison with popular counting datasets}.

\section{Experiments}
\subsection{Experimental Setup}
\textbf{Dataset.}
We evaluate TFCounter on two general object counting datasets, FSC147 and CARPK, and further study its generalizability on the proposed BIKE-1000. FSC147 contains 6135 images spanning 147 object categories, with a test subset of 1190 images from 29 categories. CARPK includes 1448 images documenting around 90,000 cars from a drone's perspective, with 459 images dedicated to testing. The BIKE-1000 dataset, with its complete set of 1000 images, serves to estimate our model's performance in a novel domain.
\textbf{ Metrics.}
To assess the accuracy of our method, we utilize Mean Absolute Error (MAE) and Root Mean Squared Error (RMSE) as metrics, both of which are established standards for object counting tasks \cite{ma2019bayesian,ranjan2018iterative}. These metrics are defined as: $MAE = \frac{1}{n} \sum_{i=1}^{n} |c_i - \hat{c}_i|$; $RMSE = \sqrt{\frac{1}{n} \sum_{i=1}^{n} (c_i - \hat{c}_i)^2}$, where $n$ denotes the number of test images, while $c_i$ and $\hat{c}_i$ represent the actual and predicted object counts, respectively.
\textbf{ Implementation.}
In the weighted fusion process of foreground and background similarity maps, we adjust $\lambda$ to 0.5 for FSC147 and to 0.7 for CARPK and BIKE-1000. Moreover, to prevent small objects from being omitted by excessive background fusion, $\lambda$ is set to 0 when the foreground regions are more than 50\%.

\subsection{State-of-the-art Comparison}
We compare our model to competitive baselines: (1)GMN\cite{lu2019class}, (2)CFOCNet+\cite{yang2021class}, (3)FamNet+\cite{ranjan2021learning}, (4)Counting-DETR\cite{nguyen2022few}, (5)BMNet+\cite{shi2022represent}, (6)SAFECount\cite{you2023few}, (7)SPDCN\cite{lin2022scale}, (8)CounTR\cite{liu2022countr}, (9)LOCA\cite{djukic2023low}, (10)Zero-shot Object Counting\cite{xu2023zero}, (11)CLIP-Count\cite{jiang2023clip}, (12)CounTX\cite{amini2023open}, (13)GroundingDINO\cite{liu2023grounding}, (14)SAM\cite{kirillov2023segment}, (15)SAM-Free\cite{shi2023training}.

\begin{table}[t!]
\centering
\setlength{\tabcolsep}{2pt} 
\renewcommand{\arraystretch}{1.1} 
\captionsetup{skip=10pt}
\caption{Quantitative comparison on FSC147 and CARPK.}
\label{tab:Quantitative comparison on FSC147 and CARPK}
\begin{tabular}{c|cccccc}
\hline
                                  &                                     &                                   & \multicolumn{2}{c}{\textbf{FSC147}}                          & \multicolumn{2}{c}{\textbf{CARPK}}                          \\ \cline{4-7} 
\multirow{-2}{*}{\textbf{Method}} & \multirow{-2}{*}{\textbf{Training}} & \multirow{-2}{*}{\textbf{Prompt}} & \textbf{MAE}                 & \textbf{RMSE}                 & \textbf{MAE}                 & \textbf{RMSE}                \\ \hline
GMN                            & Yes                                 & box                             & 26.52                        & 124.57                         & —                        & —                        \\
CFOCNet+                            & Yes                                 & box                             &  22.10                        & 112.71                         & —                        & —                        \\
FamNet+                            & Yes                                 & box                             & 22.08                        & 99.54                         & 28.84                        & 44.47                        \\
Counting-DETR                     & Yes                                 & box                             & 16.79                        & 123.56                        & —                            & —                            \\
BMNet+                            & Yes                                 & box                             & 14.62                        & 91.83                         & 10.44                        & 13.77                        \\
SAFECount                         & Yes                                 & box                             & 14.32                        & 85.54                         & 16.66                        & 24.08                        \\
SPDCN                             & Yes                                 & box                             & 13.51                        & 96.80                         & 18.15                        & 21.61                        \\
CounTR                            & Yes                                 & box                             & 11.95                        & 91.23                         & —                            & —                            \\
LOCA                              & Yes                                 & box                             & {\color[HTML]{009901} 10.79} & {\color[HTML]{009901} 56.97}  & {\color[HTML]{009901} 9.97}  & {\color[HTML]{009901} 12.51} \\ \hline
Zero-shot Object Counting         & Yes                                 & text                              & 22.09                        & 115.17                        & —                            & —                            \\
CLIP-Count                        & Yes                                 & text                              & 17.78                        & 106.62                        & 11.96                        & 16.61                        \\
CounTX                            & Yes                                 & text                              & {\color[HTML]{3531FF} 15.88} & {\color[HTML]{3531FF} 106.29} & {\color[HTML]{3531FF} 11.64} & {\color[HTML]{3531FF} 14.85} \\ \hline
GroundingDINO                     & No                                  & text                              & 59.23                        & 159.28                        & 27.72                        & 51.49                        \\
SAM                               & No                                  & N.A.                              & 42.48                        & 137.50                        & 16.97                        & 20.57                        \\
SAM-Free                              & No                                  & box                             & 19.95                        & 132.16                        & 10.97                        & 14.24                        \\
TFCounter(Ours)& No                                  & box                             & {\color[HTML]{FE0000} 18.56} & {\color[HTML]{FE0000} 130.59} & {\color[HTML]{FE0000} 9.71}  & {\color[HTML]{FE0000} 12.44} \\ \hline
\end{tabular}
\end{table}

\begin{table}[t!]
\centering
\captionsetup{skip=10pt}
\caption{Quantitative comparison on BIKE-1000.}
\label{tab:Quantitative comparison on BIKE-1000}
\setlength{\tabcolsep}{7pt} 
\renewcommand{\arraystretch}{1.1} 
\begin{tabular}{c|cccc}
\hline
\textbf{Method} & \textbf{Training} & \textbf{Prompt} & \textbf{MAE}                & \textbf{RMSE}                \\ \hline
GroundingDINO   & No                & text            & {\color[HTML]{333333} 7.95} & {\color[HTML]{333333} 12.37} \\
SAM-Free            & No                & box           & {\color[HTML]{333333} 7.43} & {\color[HTML]{333333} 10.07} \\
TFCounter(Ours)    & No                & box           & {\color[HTML]{FE0000} 6.59} & {\color[HTML]{FE0000} 10.01} \\ \hline
\end{tabular}
\end{table}

\begin{figure}[t!]
\centering
\includegraphics[height=12cm]{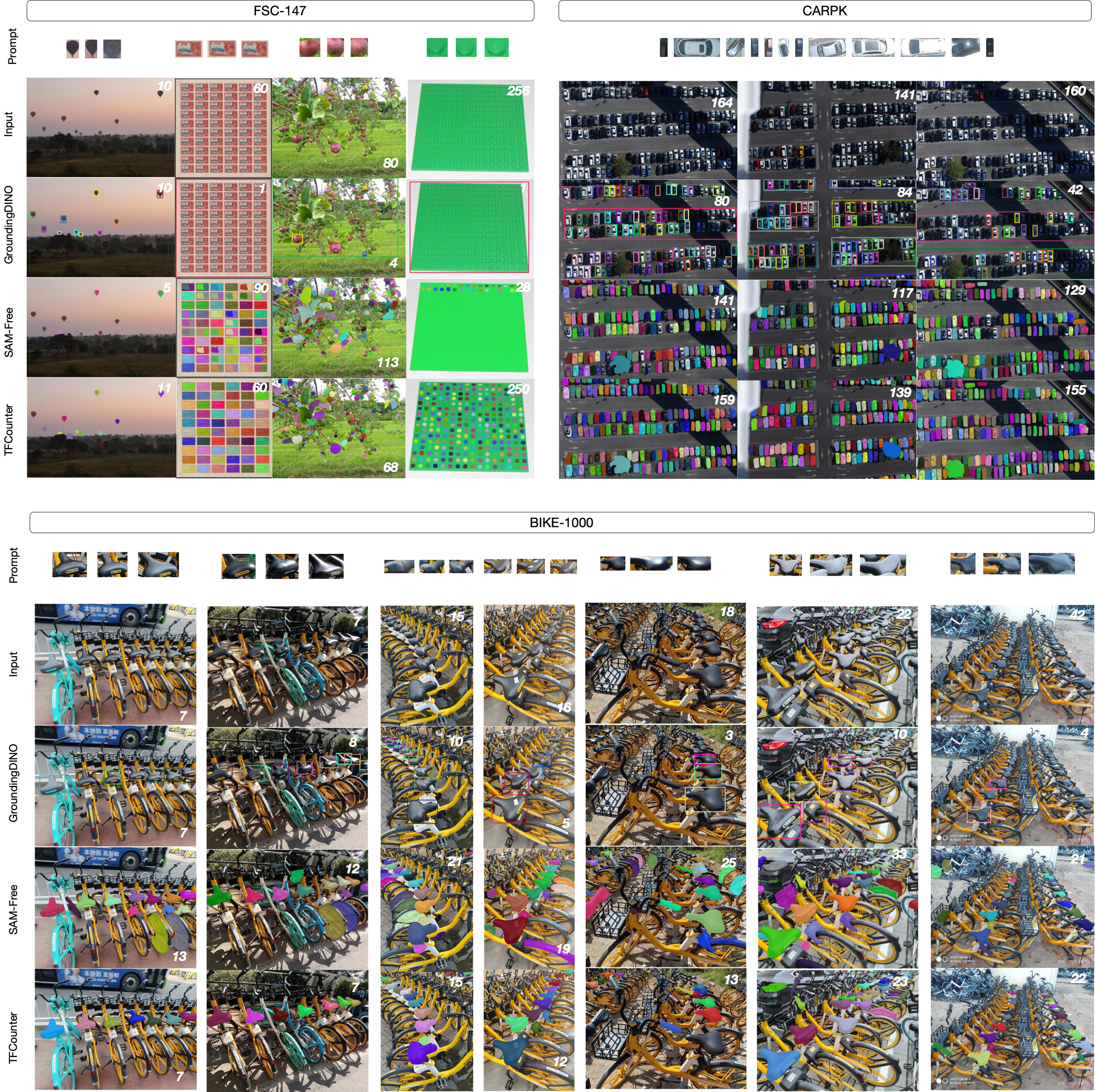}
\caption{Qualitative comparison on FSC147, CARPK, and BIKE-1000. 
The "prompt" represents box prompts for SAM-Free and TFCounter. In the CARPK dataset, we followed the approach of Ranjan et al. \cite{ranjan2021learning} by using 12 predefined examples from the training set for cross-image counting. GroundingDINO employs FSC-147-D \cite{amini2023open}, "cars", and "bike seat" as text prompts for the FSC147, CARPK, and BIKE-1000, respectively.}
\label{fig:Qualitative Results}
\end{figure}

\subsubsection{Quantitative Results on FSC147 and CARPK}
Table \ref{tab:Quantitative comparison on FSC147 and CARPK} provides a quantitative comparison of various models on the FSC147 and CARPK datasets. The top results from trained models with box prompts are highlighted in green, while the best outcomes using text prompts are in blue. The training-free model achieving the first-rate performance is indicated in red. All trained models underwent extensive training over hundreds of epochs on the FSC147 dataset. Their performance on CARPK demonstrates cross-dataset generalizability. Relatively, the assessment of training-free models on both datasets reveals zero-shot generalization capabilities. On the FSC147 dataset, TFCounter's performance outperforms that of three trained models with box prompts and another with text prompts. Specifically, TFCounter exhibits an increase in MAE of only 7.77 compared to the top-performing box-prompt model, LOCA, and a 2.68 increase compared to the leading text-prompt model, CounTX. It should be noted that LOCA and CounTX attain their results by utilizing thousands of training data and complex training methodologies. Among all training-free models, our model indicates the best performance. Compared to the present benchmark in training-free models, SAM-Free, our model reduces MAE by 1.39, lowering it from 19.95 to 18.56. Moreover, on the CARPK dataset, TFCounter attains state-of-the-art results, underscoring the robust generalization capabilities of our model.

\subsubsection{Quantitative Results on BIKE-1000}
We introduce BIKE-1000, a novel dataset with oblique perspective images of shared bicycles for object counting task to assess the cross-domain generalization capabilities of various models. In our evaluation of BIKE-1000, we benchmark the performance of GroundingDINO, SAM-Free, and our newly proposed TFCounter. GroundingDINO represents the state-of-the-art in open-set object detection, capable of object counting via aggregation of detected bounding boxes. Meanwhile, SAM-Free stands as the leading model in the research landscape of training-free, class-agnostic object counting. Table \ref{tab:Quantitative comparison on BIKE-1000} showcases the quantitative evaluation outcomes for the aforementioned models. TFCounter outperforms its counterparts, establishing a new state-of-the-art in training-free counting methods and demonstrating outstanding generalization performance on the novel BIKE-1000 dataset.

\subsubsection{Qualitative Results}
Additionally, we present a comparative analysis of the visualization results for GroundingDINO, SAM-Free, and TFCounter. Figure \ref{fig:Qualitative Results} illustrates the qualitative distinctions among the models on the FSC147, CARPK, and BIKE-1000 datasets. GroundingDINO exhibits commendable performance in counting objects of low density and demonstrates robust zero-shot generalization capabilities across multi-class objects within diverse datasets. Nevertheless, its efficacy diminishes when tasked with high-density object counting or facing significant intra-class differences. In contrast, SAM-Free surpasses GroundingDINO in high-density scenarios yet exhibits a propensity for false positives, mistaking non-target items that share a resemblance in shape or color. This issue is particularly pronounced against messy backdrops. For instance, in BIKE-1000's test images, SAM-Free frequently misidentifies bike locks and wheels—owing to their color similarity with bike seats—and handlebars due to their resemblance to shape. Moreover, SAM-Free tends to fragment a single object into multiple parts, a phenomenon conspicuously illustrated in the second visualization from the left on the FSC147 dataset. Our proposed TFCounter notably ameliorates the limitations above. It demonstrates superior performance in counting objects across both high-density and low-density scenarios. Despite facing the challenging test images from BIKE-1000, TFCounter exhibits excellent accuracy over SAM-Free by recalling fewer non-target objects and reducing the omission due to significant intra-class differences. While the MAE metric captures the absolute discrepancy between predicted counts and ground truth, it may not fully encapsulate the nuanced improvements made by TFCounter. However, these enhancements are evident in the visualization comparisons, such as in the fourth visualization from the left of the BIKE-1000 dataset.

\subsection{Ablation Studies and Analysis}

\begin{table}[t!]
\centering
\captionsetup{skip=10pt}
\caption{Analyzing the components of TFCounter.}
\label{tab:Analyzing the components of TFCounter}
\setlength{\tabcolsep}{3pt} 
\renewcommand{\arraystretch}{1.1} 
\begin{tabular}{ccc|cccc}
\hline
\multicolumn{3}{c|}{\textbf{Component}}                                                                                                                                                                    & \multicolumn{2}{c}{\textbf{FSC147}} & \multicolumn{2}{c}{\textbf{BIKE-1000}} \\ \hline
\textbf{\begin{tabular}[c]{@{}c@{}}Background\\ Similarity\end{tabular}} & \textbf{\begin{tabular}[c]{@{}c@{}}Multi-round
\\ Counting\end{tabular}} & \textbf{\begin{tabular}[c]{@{}c@{}}Residual Point\\ Prompt\end{tabular}}  & \textbf{MAE}     & \textbf{RMSE}    & \textbf{MAE}    & \textbf{RMSE}    \\ \hline
                           &                        &                        & 19.95            & 132.16           & 7.43            & 10.07            \\
$\surd$                          &                        &                        & 20.85            & 132.38           & 10.55           & 14.60            \\
                           & $\surd$                      &                        & 20.62            & 132.04           & 9.16            & 11.80            \\
                           &                        & $\surd$                      & 21.41            & 131.94           & 22.32           & 25.74            \\
$\surd$                          & $\surd$                      &                        & 20.46            & 131.79           & 10.36           & 14.43            \\
$\surd$                          &                        & $\surd$                      & 18.50            & 130.92           & 6.48            & 10.40            \\
                           & $\surd$                      & $\surd$                      & 23.22            & 132.25           & 30.16           & 35.31            \\
$\surd$                          & $\surd$                      & $\surd$                      & 18.56            & 130.59           & 6.59            & 10.01            \\ \hline
\end{tabular}
\end{table}

\begin{figure}[t!]
  \centering
\begin{minipage}[t]{0.48\textwidth}
    \includegraphics[width=\linewidth]{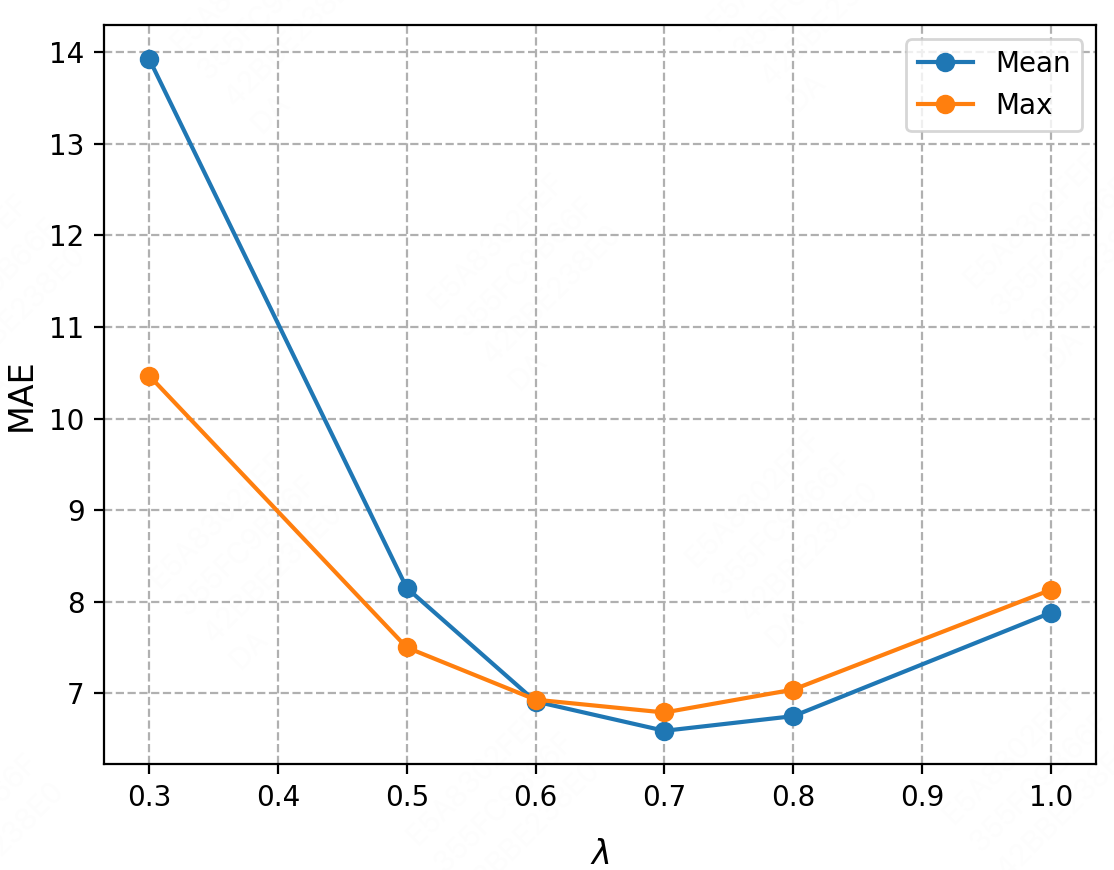}
  \end{minipage}
\begin{minipage}[t]{0.48\textwidth}
    \includegraphics[width=\linewidth]{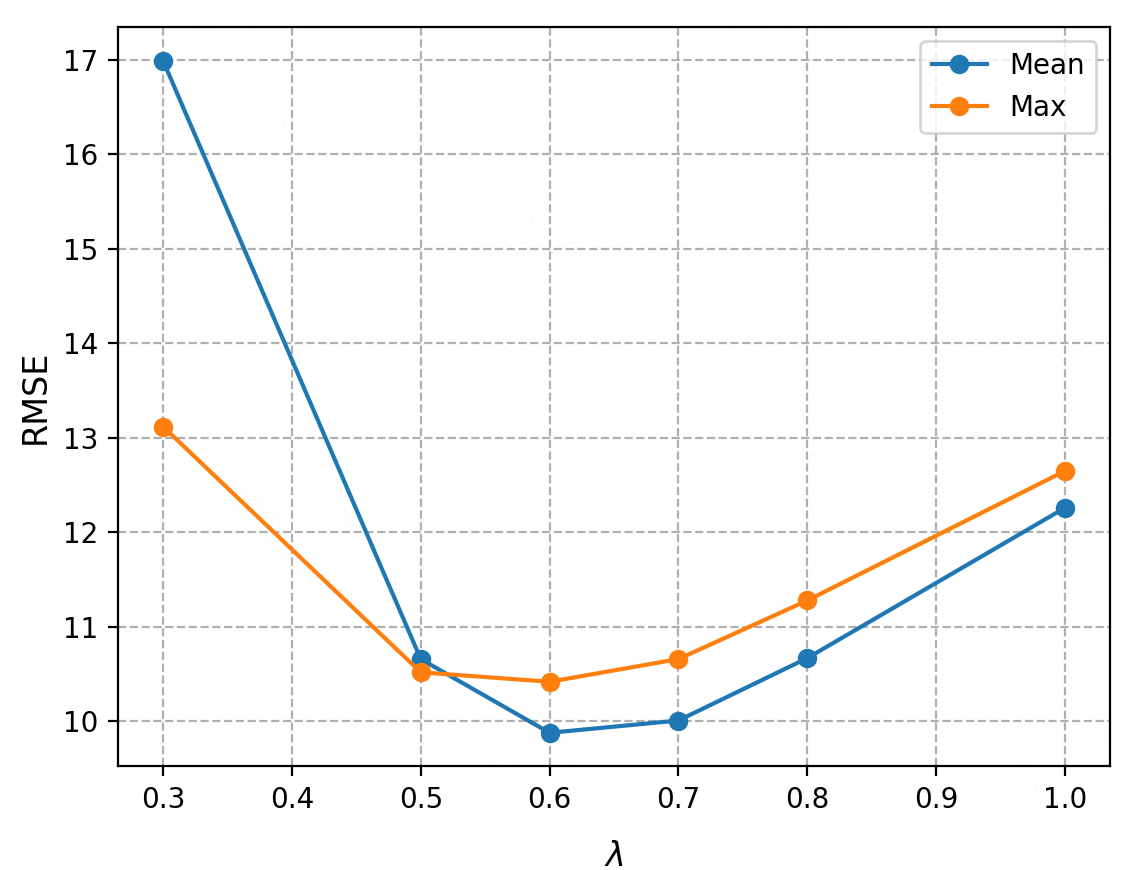}
  \end{minipage}
  \caption{Influence of hyperparameter $\lambda $ in weighted fusion process of the foreground and background similarity maps.}
  \label{fig:Influence of hyperparameter}
\end{figure}

\subsubsection{Component Analysis}
To substantiate the efficacy of each component in TFCounter
, we executed a series of ablation studies, the results of which are delineated in Table \ref{tab:Analyzing the components of TFCounter}. The first row of the table presents the SAM-Free, which offers a well-designed counting module for the general class-agnostic counting framework. 
The evaluation data presented in rows 2 to 4 of the table reveal that the isolated implementation of any single component—be it the Background Similarity, Multi-round Counting, or Residual Point Prompt—results in diminished performance. The assessments depicted in rows 5 to 7 further corroborate that the binary combinations of these components are suboptimal, with the fusion of Multi-round Counting and Residual Point Prompt being particularly ineffective. However, as demonstrated in row 8, the synergistic integration of all three components yields pleasing outcomes. It can be attributed to the distinctive functions of each component: the Background Similarity is tasked with filtering out irrelevant masks to bolster accuracy, whereas the Multi-round Counting and Residual Point Prompt are designed to broaden the recall scope, thereby encapsulating a larger array of target objects. Optimal performance is realized through the harmonious interplay of these components, as the exclusive use of the Background Similarity may lead to the inadvertent exclusion of smaller objects, and reliance solely on Multi-round Counting and Residual Point Prompt could result in the erroneous inclusion of non-target objects. Thus, the efficacy of our approach is predicated on the collaborative operation of these components.

\subsubsection{Hyperparameters Analysis}
Subsequently, we explored the influence of the hyperparameter $\lambda$ in the weighted fusion process of the foreground and background similarity maps. We experimented with two fusion methods: 1) the mean fusion, denoted as "Mean", formulated as $\frac{1}{N} \sum_{i=1}^{N} fsim_{i} + \lambda \times bsim$; and 2) the maximum fusion, denoted as "Max", formulated as $\max_{i} (fsim_{i}) + \lambda \times bsim$. Figure \ref{fig:Influence of hyperparameter} illustrates the impact of different $\lambda$ values and fusion methods on the performance of TFCounter on the BIKE-1000 dataset. The results indicate that as $\lambda$ increases, both MAE and RMSE decrease initially and then increase, with only a slight difference in the optimal point (from 0.7 to 0.6). It suggests an optimal ratio for the fusion process on the BIKE-1000 dataset, and fine-tuning this ratio for each image might yield better accuracy, which is a potential direction for future research. Moreover, when $\lambda$ is small, the "Mean" method incurs a more significant mistake, whereas the reverse is for larger $\lambda$. A possible reason is that the "Mean" method considers all exemplars, making it easier to recall non-target objects when less background fusion; the "Max" method focuses only on the most similar exemplar, resulting in more omission of small objects when excessive background fusion. Overall, both methods exhibit limitations that emphasize the need for future work to more universal fusion strategies. In this paper, the "Mean" method with $\lambda=0.7$ is adopted for the BIKE-1000 dataset.

\begin{figure}[t!]
  \centering
\begin{minipage}[t]{0.48\textwidth}
    \includegraphics[width=\linewidth]{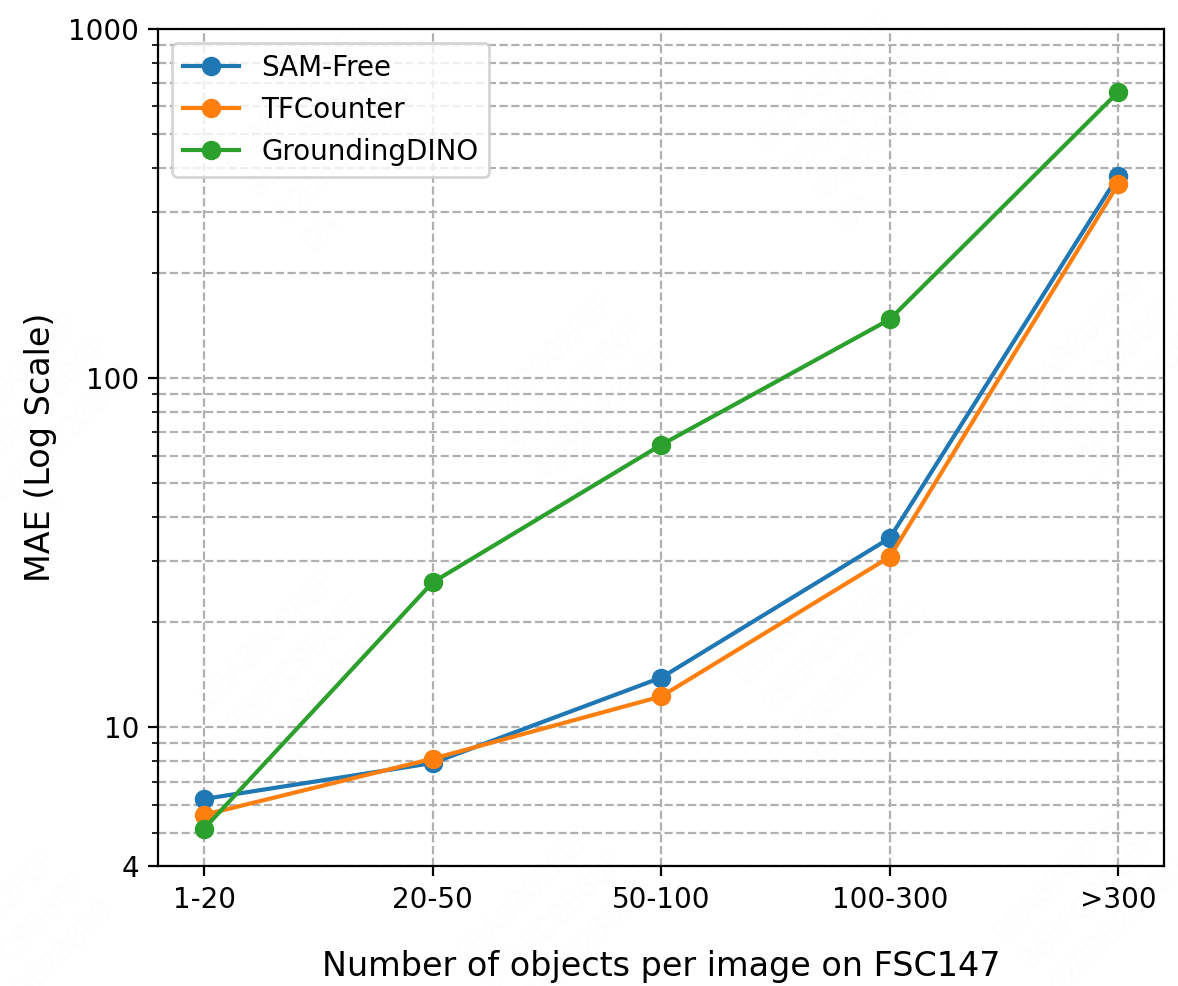}
  \end{minipage}
\begin{minipage}[t]{0.48\textwidth}
    \includegraphics[width=\linewidth]{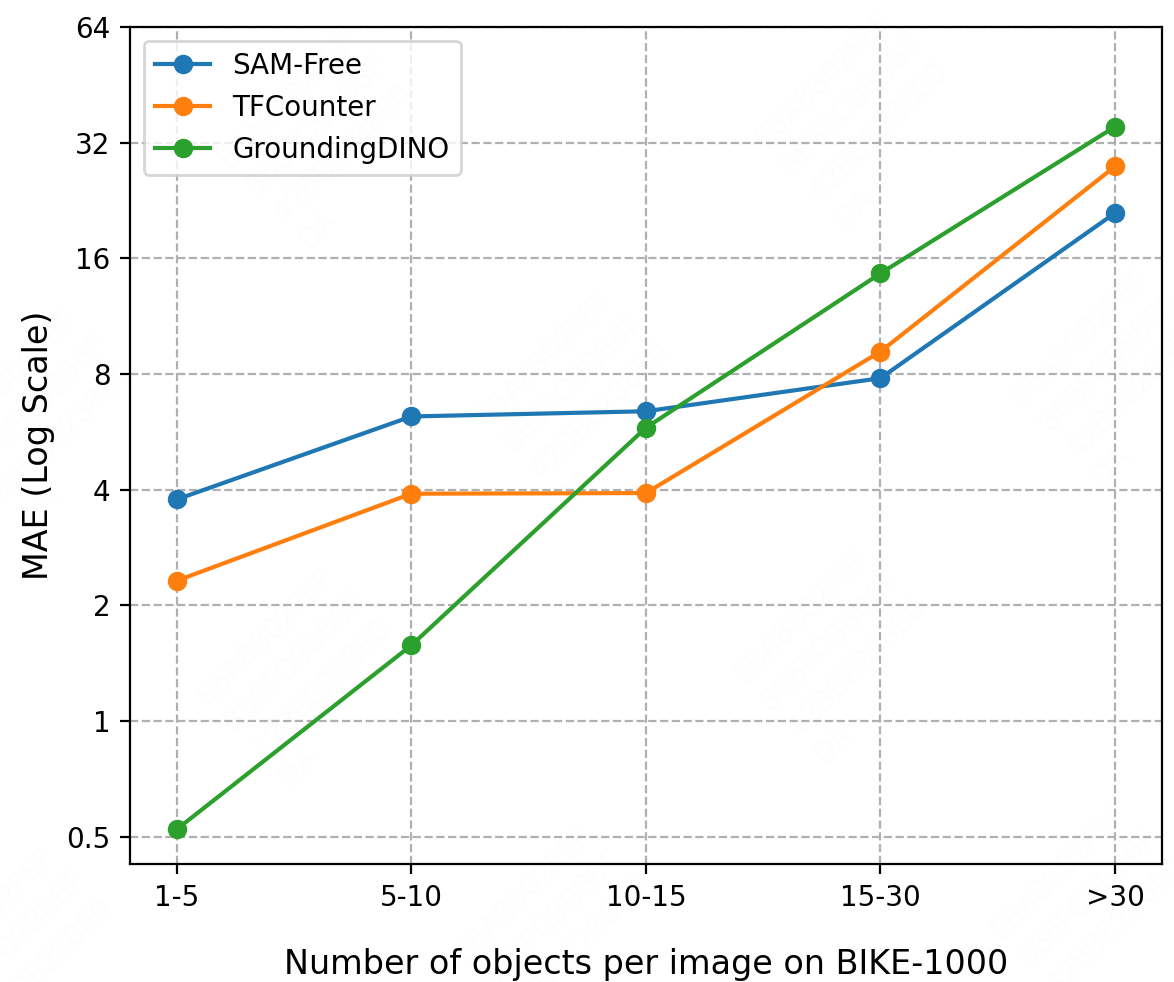}
  \end{minipage}
  \caption{Performance in different density images.}
  \label{fig:Performance in different density images}
\end{figure}

\subsubsection{Density Analysis}
Finally, we compared three training-free methods: GroundingDINO, SAM-Free, and TFCounter, in terms of their performance on test images with various densities. Figure \ref{fig:Performance in different density images} showcases MAE on the FSC147 and BIKE-1000 datasets, where the vertical axis is presented on a logarithmic scale for better visualization. GroundingDINO performs optimally on low-density test images, but its MAE increases exponentially with rising density. TFCounter demonstrates superior accuracy in medium to low-density scenarios. Meanwhile, SAM-Free reaches or slightly outperforms TFCounter in high-density test images. However, SAM-Free performs better in high-density test images partly due to recalling more non-target objects rather than accurately counting all targets, which unexpectedly brings the counting results closer to the true value. A typical example of this can be seen in the fourth visualization from the left in Figure \ref{fig:Qualitative Results} of the BIKE-1000 dataset.

\section{Limitations}
The initial version of TFCounter presents several limitations.
\textbf{ Computational Efficiency.} Our model achieves object counting through mask segmenting, multi-round counting strategy, and residual point prompt, all of which result in an increased processing time, particularly for high-density images. Future research on more lightweight large-scale segmentation models may mitigate this limitation. 
\textbf{ Segmentation Granularity.} Our model utilizes a segmentation-based approach for object counting by tallying the number of masks. However, when objects comprise multiple parts with significant color or shape differences, such as the red flesh and green calyx of a strawberry, the segmentation-based model may encounter issues with double counting due to segmenting a single object into multiple masks. One future solution direction is to explore large-scale foundation models with adaptive segmentation granularity.

\section{Conclusions}
In this paper, we explore a training-free technique for processing downstream tasks and applications in computer vision by integrating generalizable components from large-scale foundation models into task-specific frameworks. Based on it, we propose TFCounter, an innovative model capable of performing training-free category-agnostic object counting tasks using visual prompts. The originality of TFCounter stems from three core designs: a multi-round counting strategy, a dual prompt system, and a context-aware similarity module. The first two contribute to broadening the recall scope, while the latter boosts accuracy by incorporating background context. The development of TFCounter hopes to excite further thinking about how to adapt large-scale foundation models well-known for high generalizability to various downstream tasks and domain data, simultaneously maintaining superior performance. Future works include innovative visual prompts for more intuitive human-computer interactions and more adaptive designs for the similarity module to enhance performance.


%
%
\bibliographystyle{splncs04}
\bibliography{egbib}
\end{document}